\documentclass[letterpaper, 10 pt, conference]{ieeeconf}  

\IEEEoverridecommandlockouts                              

\usepackage{times}

\usepackage{soul}
\usepackage{url}
\usepackage[hidelinks]{hyperref}
\usepackage[small]{caption}
\usepackage{graphicx}
 
\usepackage{amsthm}
\usepackage{amsmath,bm}
\usepackage{booktabs}
\usepackage{algorithm}
\usepackage{algorithmic}
\usepackage{booktabs}
\usepackage{multirow}
\usepackage{booktabs}
\usepackage{threeparttable}
\usepackage{makecell}
\usepackage{graphicx}
\usepackage{subfigure} 
\usepackage{amsfonts,amssymb}
\usepackage{cite}
\usepackage{float}  
\newtheorem{theorem}{Theorem}
\newtheorem{lemma}{Lemma}

\title{\LARGE \bf
Triple-GAIL: A Multi-Modal Imitation Learning Framework with Generative Adversarial Nets
}

\author{Cong Fei$^{1,2\dag}$, Bin Wang$^{1\dag}$, Yuzheng Zhuang$^{1\dag}$,	Zongzhang Zhang$^3$, Jianye Hao$^{1}$, \\Hongbo Zhang$^{1}$, Xuewu Ji$^2$ and Wulong Liu$^{1}$ 
\thanks{$^{\dag}$The first three authors contribute equally. C. Fei conducted this work during the internship in Huawei Noah's Ark Lab. B. Wang is the corresponding author} 
\thanks{$^{1}$B. Wang, Z. Zhuang, J. Hao, H. Zhang and W. Liu are all with Huawei Noah's Ark Lab, Beijing, 100085, China
	{\tt\small \{wangbin158, zhuangyuzheng, haojianye, zhanghongbo888, liuwulong\}@huawei.com}}%
\thanks{$^{2}$C. Fei and X. Ji are with Tsinghua University, Beijing, 100084, China
        {\tt\small feic18@mails.tsinghua.edu.cn, jixw@mail.tsinghua.edu.cn}}%
\thanks{$^3$Z. Zhang is with Nanjing University, Nanjing, 210023, China
	{\tt\small zhangzongzhang@gmail.com}}%
}

\begin{document}

\maketitle
\thispagestyle{empty}
\pagestyle{empty}

\begin{abstract}

Generative adversarial imitation learning (GAIL) has shown promising results by taking advantage of generative adversarial nets, especially in the field of robot learning. However, the requirement of isolated single modal demonstrations limits the scalability of the approach to real world scenarios such as autonomous vehicles' demand for a proper understanding of human drivers' behavior. In this paper, we propose a novel multi-modal GAIL framework, named Triple-GAIL, that is able to learn skill selection and imitation jointly from both expert demonstrations and continuously generated experiences with data augmentation purpose by introducing an auxiliary skill selector. We provide theoretical guarantees on the convergence to optima for both of the generator and the selector respectively. Experiments on real driver trajectories and real-time strategy game datasets demonstrate that Triple-GAIL can better fit multi-modal behaviors close to the demonstrators and outperforms state-of-the-art methods.

\end{abstract}

\section{INTRODUCTION}

Imitation learning aims to mimic expert behavior directly from human demonstrations, without designing explicit reward signal as reinforcement learning (RL) \cite{abbeel2004apprenticeship,fang2019survey}, and has made achievements in a variety of tasks. 
Recent work in imitation learning, especially generative adversarial imitation learning (GAIL) \cite{ho2016generative}, optimizes a policy directly from expert demonstrations without estimating the corresponding reward function, and overcomes compounding errors caused by behavioral cloning (BC) \cite{reddy2019sqil} as well as reduces the computational burden of inverse reinforcement learning (IRL) \cite{wulfmeier2015maximum,pflueger2019rover}. Existing imitation learning methods, including GAIL, mostly focus on reconstructing expert behavior based on the assumption of single modality. However, most of real world demonstrations have multiple modalities with various skills and habits. For example, there are three distinct intentions in a driving task: lane-change left, lane keeping and lane-change right. In that case, imitation learning algorithms like GAIL will cause the mode collapse problem due to the inability of discovering and distinguishing mode variation in expert demonstrations. Besides, much of real world tasks like aforementioned driving task need to select behavior mode based on current situation for decision-making adaptively (i.e., human drivers will determine whether to change lanes based on traffic conditions) instead of specified manually.

Some extensions of GAIL have been proposed to deal with multi-modal tasks. \cite{li2017infogail,kuefler2018burn,wang2017robust} learn latent codes in an unsupervised manner and recover multi-modal policies from unlabeled demonstrations, which need random sampling of latent codes. \cite{merel2017learning,lin2018acgail} reconstruct modal information directly from expert demonstration labels or add an auxiliary classifier to assist the adversary, involving a supervised learning process. However, most of the extensions only focus on learning to distinguish different skills with random sampling of skill labels, thus they are not able to deal with those real world scenarios which require adaptive skill selection conditioned on environmental situations. This motivates our research.

In this paper we propose a new approach learning to select skill labels and imitate multi-modal policy simultaneously. The algorithm, called Triple-GAIL, is an extension of GAIL for distinguishing multiple modalities accurately and efficiently enhancing the performance on label-conditional imitation learning tasks. In particular, the contributions of this paper are as follows: (i) Similar to Triple-GAN \cite{chongxuan2017triple}, we propose a novel adversarial game framework which extends the original GAIL with an auxiliary selector. The selector and the generator in Triple-GAIL characterize the conditional distribution given the state-action pairs and state-label pairs while the discriminator distinguishes whether a state-action-label pair comes from expert demonstrations or not. (ii) Both of the generator and the selector have been proved to converge to their own optima respectively with compatible utilities, which means Triple-GAIL can learn a good skill selector and a conditional generator simultaneously. (iii) We apply our algorithm in a driving task and a real-time strategy (RTS) game. Experimental results demonstrate that Triple-GAIL can distinguish multiple modalities clearly as well as enhance the performance on label-conditional imitation learning tasks.

\section{BACKGROUND AND RELATED WORK}
\subsection{Generative Adversarial Imitation Learning}
GAIL is a promising imitation learning method based on generative adversarial nets (GANs) \cite{goodfellow2014generative}. In GAIL, the generator serves as a policy to imitate expert behavior by matching the state-action $(s,a)$ distribution of demonstrations, while the discriminator plays a role of surrogate reward to measure the similarity between the generated data and demonstration data. GAIL directly optimizes the policy without solving the reward function in IRL. The objective of GAIL is formulated as the min-max form:
\begin{align}
\begin{split}
\min_\pi \max_{D\in(0,1)}&\mathbb{E}_\pi\left[\log D(s,a)\right]\\
+&\mathbb{E}_{\pi_E}\left[\log\left(1-D(s,a)\right)\right]-\lambda H(\pi)
\end{split}
\end{align}
where $\pi$ and $D$ are the generator (policy) and the discriminator respectively. The casual entropy $H(\pi)$ serves as a regularization term of policy together with hyper-parameter $\lambda$. In practice, trust region policy optimization (TRPO) \cite{schulman2015trust} is used to update the policy $\pi$ with the surrogate reward function: $r=-\log D(s,a)$.

\subsection{Multi-modal Imitation Learning Algorithms}
There have been some extensions of GAIL to address multi-modal tasks. One typical way is to distinguish modal information in an unsupervised manner. InfoGAIL \cite{li2017infogail} infers latent codes by maximizing the mutual information
between latent variables and observed state-action pairs. Burn-InfoGAIL \cite{kuefler2018burn} uses the maximum mutual information from the perspective of Bayesian inference to draw modal variables from burn-in demonstrations. VAE-GAIL \cite{wang2017robust} introduces a variational autoencoder to infer modal variable, which allows for smoothing policy interpolation. The above algorithms can learn multi-modal policies from unlabeled demonstrations. However, due to lack of labels in demonstrations, these algorithms tend to distinguish latent labels without considering semantic information or task context.

Another direction focuses on labeled data in expert demonstrations. In CGAIL \cite{merel2017learning}, the modal labels are directly sent to the generator and the discriminator, which are conditioned onto the label itself. ACGAIL \cite{lin2018acgail} introduces an auxiliary classifier to reconstruct the modal information specially while the discriminator is only responsible for distinguishing whether the input sample $(s,a)$ is from the demonstrations. In ACGAIL the classifier cooperates with the discriminator by sharing parameters, both of which provide adversarial loss to the generator. 

Note that that above methods mainly leverage random sampling of latent labels from known prior distribution to distinguish multiple modalities. Once trained,  model outputs the corresponding actions based on the manually specified labels. However, in this paper we focus on dealing with those tasks which require adaptive skill mode selection according to environmental situations. Moreover, we are interested in labeled expert demonstrations with multiple modalities. Different from existing works, the proposed Triple-GAIL is able to learn skill selection and imitation jointly from both expert demonstrations and continuously generated experiences.

\section{METHOD}
Suppose we can get a mixed set of labeled demonstrations with multiple expert modalities. In this paper we propose to learn one policy simultaneously from multiple expert demonstrations. More specifically, the expert policy including multiple skill labels is presented as $\pi_{E}=\left\{\pi_{E_1},...,\pi_{E_k}\right\}$, which is determined by $p(\pi|c)$, where $c$ is the skill label. In order to select skill labels from current environmental observations adaptively instead of specifying manually and then reconstruct multi-modal policy simultaneously
, a novel adversarial imitation framework is introduced as follows.
\begin{figure}[tb]
	\centering
	\includegraphics[width=3.3 in]{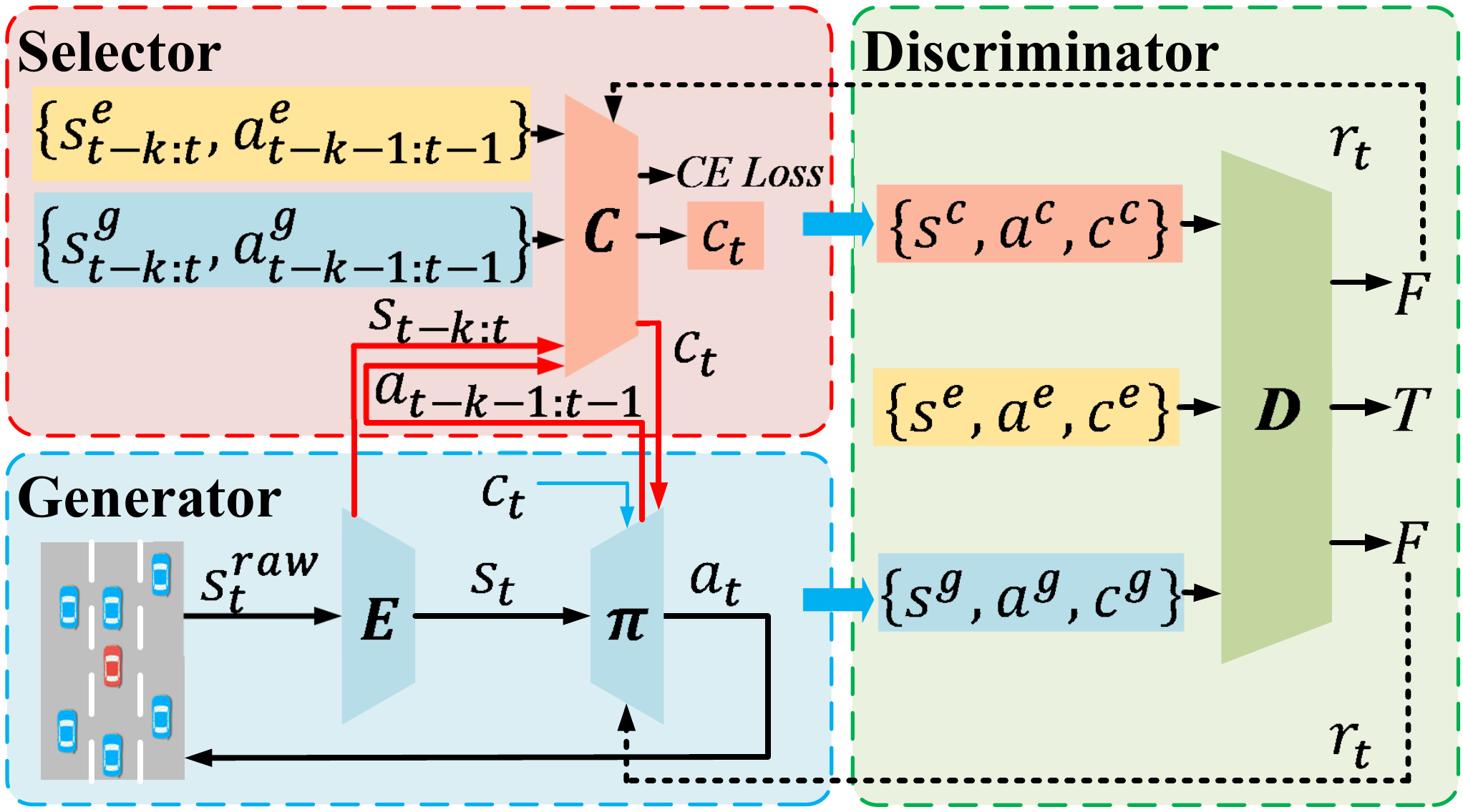}
	\caption{The overall structure of Triple-GAIL. Label $c$ is drawn from expert data (blue line) in the training process, while inferred from state-action pairs in application (red line), which characterizes the conditional distribution $p_{C_\alpha}$ (red block). The generator characterizes the conditional distribution $p_{\pi_{\theta}}$ (blue block). All these data sequences are sent to the discriminator together with expert data $p_{\pi_E}$. $t-k:t$ means time sequences. $\bm E$ denotes an encoder.}
	\label{fig:stru}
\end{figure}
\subsection{Triple-GAIL Framework}
Triple-GAIL consists of three main components represented by neural networks as shown in Figure \ref{fig:stru}: a selector $C_{\alpha}$ parameterized by $\alpha$, which produces skill labels given states and actions; a generator $\pi_{\theta}$ parameterized by $\theta$, which acts as a corresponding policy conditioned on states and skill labels; and a discriminator $D_\psi$ parameterized by $\psi$, which distinguishes whether the state-action-label pairs come from expert demonstrations or not. The joint distribution of state-action-label pairs in the setting of Triple-GAIL can be defined in two directions: 
$\pi_{\theta}$ approximately characterizes the conditional distribution $p_{\pi_{\theta}}(a|s,c)$ given $s$ and $c$, while $C_{\alpha}$ characterizes the conditional distribution $p_{C_{\alpha}}(c|s,a)$ given $s$ and $a$. We make the mild assumption that $p(s,c)$ and $p(s,a)$ can be obtained from the demonstrations and generated data respectively, then the two distributions are defined as follows:
\begin{align}
p_{\pi_{\theta}}(s,a,c)&=p(s,c)p_{\pi_{\theta}}(a|s,c)\label{3}\\
p_{C_{\alpha}}(s,a,c)&=p(s,a)p_{C_{\alpha}}(c|s,a)
\label{4}
\end{align}
where $p_{\pi_{\theta}}(s,a,c)$ and $p_{C_{\alpha}}(s,a,c)$ are the joint distributions defined by $\pi_{\theta}$ and $C_{\alpha}$ respectively, and $p_{\pi_{E}}(s,a,c)$ denotes the expert distribution. We expect to achieve an equilibrium that both $p_{\pi_{\theta}}(s,a,c)$ and $p_{C_{\alpha}}(s,a,c)$ converge to the expert data distribution.

In this game, we can draw skill labels $c$ from expert demonstrations and then produces actions $a$ conditioned on the current states and labels following Eqn. \eqref{3}, which generates pseudo joint pairs $(s^g,a^g,c^g)$. Similarly, the selector provides pseudo skill labels given the current states and last actions generated from interaction, hence pseudo joint pairs $(s^c,a^c,c^c)$ can also be sampled following Eqn. \eqref{4}. Besides, real joint pairs $(s^e,a^e,c^e)$ from expert demonstrations are provided as positive samples. Then, the joint pairs sampled from $p_{\pi_{\theta}}(s,a,c)$, $p_{C_{\alpha}}(s,a,c)$ and $p_{\pi_{E}}(s,a,c)$ are all sent to the discriminator $D_\psi$ for judgement. Note that different from Triple-GAN, which introduces a classifier for label classification in semi-supervised learning, the selector in Triple-GAIL in essence plays the role of skill selection under current circumstance, not just for distinguishing skill labels.

In such a three-player game, the generator and the selector work cooperatively against the discriminator. The adversarial loss, in turn, guides the generator as well as refines the selector, resulting in a multi-modal policy. In analogy with GAIL, the framework of Triple-GAIL is first formulated as a game to minimize $\alpha$, $\theta$ and maximize $\psi$: 
\begin{equation}
\begin{split}
\min _{\alpha,\theta}\max _{\psi}&\mathbb{E}_{\pi_{E}}\left[\log \left(1-D_{\psi}(s, a, c)\right)\right]\\
+&\omega \mathbb{E}_{\pi_{\theta}}\left[\log D_{\psi}(s, a, c)\right]\\ 
+&(1-\omega) \mathbb{E}_{C_{\alpha}}\left[\log D_{\psi}(s, a, c)\right]-\lambda_{H} H\left(\pi_{\theta}\right) 
\end{split}
\label{5}
\end{equation}
where $\mathbb{E}_{\pi_{E}}$, $\mathbb{E}_{\pi_{\theta}}$ and $\mathbb{E}_{C_{\alpha}}$ denote $\mathbb{E}_{(s,c,a)\sim p_{\pi_E}(s,c,a)}$, $\mathbb{E}_{(s,c,a)\sim p_{\pi_{\theta}}(s,c,a)}$, $\mathbb{E}_{(s,c,a)\sim p_{C_\alpha}(s,c,a)}$ respectively, $\omega\in(0,1)$ is a hyper-parameter that balances the weights of policy generation and skill selection, and $H\left(\pi_{\theta}\right)$ is the policy casual entropy defined as $\mathbb{E}_{\pi_{\theta}}\left[-\log \pi_{\theta}(a | s, a)\right]$ with hyper-parameter $\lambda_H>0$. 

Since the optimal solution should be obtained if and only if the pseudo distributions defined by the generator and the selector both converge to the distribution of expert, we introduce two cross-entropy terms $R_E$ and $R_G$ (See in Sec. \ref{proofs}), and define the objective function of this three-player game as: 
\begin{equation}
\begin{aligned} 
\min _{\alpha, \theta} \max _{\psi} & \mathbb{E}_{\pi_{E}}\left[\log \left(1-D_{\psi}(s, a, c)\right)\right]\\
&+\omega \mathbb{E}_{\pi_{\theta}}\left[\log D_{\psi}(s, a, c)\right] \\ 
&+(1-\omega) \mathbb{E}_{C_{\alpha}}\left[\log D_{\psi}(s, a, c)\right] \\ 
&+\lambda_{E} R_{E}+\lambda_{G} R_{G}-\lambda_{H} H\left(\pi_{\theta}\right) 
\end{aligned}
\label{10}
\end{equation}
where $\lambda_E$ and $\lambda_G$ weigh the relative importance of two supervised loss. 
\subsection{Objective Design and Theoretical Analysis}
\label{proofs}
This section provides the formal theoretical analysis of Triple-GAIL. Following the similar proof framework of Triple-GAN, first the theoretical prediction form of the optimal discriminator can be derived as in Lemma \ref{lemma1}:
\begin{lemma}
	For any fixed generator and selector, the optimal form of the discriminator is denoted as:
	\begin{equation}
	D_{\psi^{*}}=\frac{\omega p_{\pi_{\theta}}+(1-\omega) p_{C_{\alpha}}}{p_{\pi_{E}}+\omega p_{\pi_{\theta}}+(1-\omega) p_{C_{\alpha}}}=\frac{p_{\omega}}{p_{\pi_{E}}+p_{\omega}}
	\label{6}
	\end{equation}
	where $p_{\pi_{E}}$, $p_{\pi_{\theta}}$ and $p_{C_{\alpha}}$ denote $p_{\pi_{E}}(s, a, c)$, $p_{\pi_{\theta}}(s, a, c)$ and $p_{C_{\alpha}}(s, a, c)$ respectively, and $p_{\omega}$ is defined as $\omega p_{\pi_{\theta}}+(1-\omega) p_{C_{\alpha}}$.
	\label{lemma1}
\end{lemma}

Given the optimal ${D_{\psi {\rm{*}}}}$ and omit policy entropy term, we can derive the equilibrium conditions and properties.

\begin{lemma}
	The min-max game in Eqn. \eqref{5} can achieve the multiple equilibrium that $p_{\pi_{E}}=p_{\omega}$, where variable $\omega$ is a mixing coefficient between ${p_{{\pi _\theta }}}\left( {s,a,c} \right)$ and ${p_{{C_\alpha }}}\left( {s,a,c} \right)$.
	\label{lemma2}
\end{lemma}

Since $\omega$ is a variable, Lemma \ref{lemma2} only proves a mixed distribution of ${p_{{\pi _\theta }}}\left( {s,a,c} \right)$ and ${p_{{C_\alpha }}}\left( {s,a,c} \right)$ converges to the true distribution of expert but fails to guarantee each of them converges to ${p_{{\pi _E}}}\left( {s,a,c} \right)$, that is ${p_{{\pi _\theta }}}\left( {s,a,c} \right) = {p_{{C_\alpha }}}\left( {s,a,c} \right) = {p_{{\pi _E}}}\left( {s,a,c} \right)$. To address this problem, we introduce two cross-entropy terms $R_E$ and $R_G$ as follows: 
\begin{align}
R_{E}&=\mathbb{E}_{\pi_{E}}\left[-\log p_{C_{\alpha}}(c | s, a)\right]\nonumber\\
&\approx -\frac{1}{N} \sum_{i=0}^{N} \frac{1}{T} \sum_{t=1}^{T} c_{i,t}^{e} \log p_{C_{\alpha}}\left(c_{i,t}^{c} | s_{i,t}^{e}, a_{i,t-1}^{e}\right)
\label{8}
\end{align}
\begin{align}
R_{G}&=\mathbb{E}_{\pi_{\theta}}\left[-\log p_{C_{\alpha}}(c | s, a)\right]\nonumber\\
&\approx -\frac{1}{N} \sum_{i=0}^{N} \frac{1}{T} \sum_{t=1}^{T} c_{i,t}^{g} \log p_{C_{\alpha}}\left(c_{i,t}^{c} | s_{i,t}^{g}, a_{i,t-1}^{g}\right)
\label{9}
\end{align}
where the superscripts $e$, $g$ and $c$ indicate the samples provided by the discriminator, the generator and the selector respectively. The subscript indicates the timestep. Consider that labels are drawn from expert demonstrations in the training phase, $c_t^e=c_t^g$. $R_E$ is the standard supervised loss ensuring that the selector  converges to expert distribution. $R_G$ is essentially the divergence between the pseudo distribution $p_{C_{\alpha}}(s,c,a)$ and generated distribution $p_{\pi_{\theta}}(s,c,a)$. And this optimizes the selector using the generated data from interaction, which can be viewed as data augmentation for the selector. A learning rate schedule is also introduced for $R_G$ to boost training performance. The advantage of $R_G$ is shown in Sec. 4. By combining Eqn. \eqref{8} and Eqn. \eqref{9} to the initial objective Eqn. \eqref{5}, we can obtain the final form of objective given in Eqn. \eqref{10}:

\begin{theorem}
	Eqn. \eqref{10} ensures the existence and uniqueness of the global equilibrium, which is achieved if and only if ${p_{{\pi _\theta }}}\left( {s,a,c} \right) = {p_{{C_\alpha }}}\left( {s,a,c} \right) = {p_{{\pi _E}}}\left( {s,a,c} \right)$.
	\label{theorem1}
\end{theorem}
Following Theorem \ref{theorem1}, we can guarantee both the generator and the selector can converge to their optima respectively.

\begin{algorithm*}[htp]
	\caption{The Training Procedure of Triple-GAIL}
	\label{alg:algorithm}
	\textbf{Input}: The multi-intention trajectories of expert $\tau_E$;\\ \textbf{Parameter}: The initial parameters $\theta_0$, $\alpha_0$ and $\psi_0$
	\begin{algorithmic}[1]
		\FOR{$i=0,1,2,\cdots$}
		\FOR{$j=0,1,2,\cdots,N$}
		\STATE Reset environments by the demonstration episodes with fixed label $c_j$;\
		\STATE Run policy ${\pi _\theta }\left( { \cdot |{c_j}} \right)$ to sample trajectories: ${\tau _{{c_j}}} = \left( {{s_0},{a_0},{s_1},{a_1},...{s_{T_j}},{a_{T_j}}|{c_j}} \right)$\
		\ENDFOR
		\STATE Update the parameters of $\pi_\theta$ via TRPO with rewards: ${r_{{t_j}}} = -\log {{D_\psi }\left( {{s_{{t_j}}},{a_{{t_j}}},{c_j}} \right)}$\
		\STATE Update the parameters of $D_\psi$ by gradient ascending with respect to:\\
		\begin{align}
		\nabla_{\psi}\frac{1}{{{N_e}}}\sum\limits_{n = 1}^{{N_e}}{\log(1{\rm{ - }}{D_\psi }\left( {s_n^e,a_n^e,c_n^e} \right))}  + \frac{1}{N}\sum\limits_{j = 1}^N {\left[ {\frac{\omega }{{{T_j}}}\sum\limits_{t = 1}^{{T_j}} {\log{D_\psi }\left( {s_{t}^g,a_{t}^g,c_j^g} \right) + \frac{{1 - \omega }}{{{T_j}}}\sum\limits_{t = 1}^{{T_j}} {\log {{D_\psi }\left( {s_t^c,a_t^c,c_j^c} \right)} } } } \right]}
		\label{algo:d}
		\end{align}\
		\STATE Update the parameters of $C_\alpha$ by gradient descending with respect to:\\
		\begin{align}
		\nabla_{\alpha}\frac{1}{N}\sum\limits_{j = 1}^N {\left[ {\frac{{1{\rm{ - }}\omega }}{{{T_j}}}\sum\limits_{t = 1}^{{T_j}} {\log {{D_\psi }\left( {s_t^c,a_t^c,c_j^c} \right)}} {\rm{ - }}\frac{{{\lambda _E}}}{{{T_j}}}\sum\limits_{t = 1}^{{T_j}} {c_j^e\log {p_{{C_\alpha }}}\left( {c_{{t}}^c{\rm{|}}s_{{t}}^e,a_{{t} - 1}^e} \right)} {\rm{ - }}\frac{{{\lambda _G}}}{{{T_j}}}\sum\limits_{t = 1}^{{T_j}} {c_j^e\log {p_{{C_\alpha }}}\left( {c_{{t}}^c{\rm{|}}s_{{t}}^g,a_{{t} - 1}^g} \right)} } \right]}
		\label{algo:c}
		\end{align}\
		\ENDFOR
	\end{algorithmic}
\end{algorithm*}

The whole training procedure of Triple-GAIL is summarized in Algorithm \ref{alg:algorithm}. Triple-GAIL has three models consisting of four neural networks: The generator $\pi_\theta$ consists of a policy network and a value network. The selector and the discriminator are characterized by the selector network $C_\alpha$ and the discriminator network $D_\psi$, respectively. Firstly, we reset the environment by the labeled episodes, namely, each episode has a fixed true label. The agent runs the policy with these labels and gathers the generated data. The generated data are then sent to the selector to produce skill labels. The data generated by generator and selector are all sent to the discriminator as the pseudo data while the demonstration data is served as true data. Then the discriminator network is updated by ascending the gradient with above sampled data with \eqref{algo:d} while the selector network is updated by descending the gradient with \eqref{algo:c}. Our policy network and value network are updated by TRPO.
Once trained, the selector adaptively generates skill label based on state-action pairs, and the skill label is input to the generator to produce corresponding actions.

\section{EXPERIMENTS}
Note that we tend to solve real world tasks which need to adaptively select skill mode and guide decision-making based on current environmental situations. So we demonstrate the performance of our method on two typical tasks, both of which drive their policies with multiple explicit skills under specific circumstances. We first apply it to a driving task where the agent adaptively selects whether to change lanes based on the highway traffic conditions and imitates corresponding driving behaviors. Then, we extend it to an RTS game, where the agent needs to choose its skills considering the enemy's tactical intentions and take corresponding strategies. For both tasks, our algorithm is evaluated against three baselines: BC, GAIL and CGAIL. Note that InfoGAIL and ACGAIL need to manually specify skill labels by experts, which beyond our comparison domain. We modify original CGAIL by adding a classifier with the same structure of selector in Triple-GAIL. This classifier is trained by supervised learning and the parameters are fixed. 
\subsection{Experimental Setup}
For the driving task, we follow the state and action representations as in \cite{henaff2019model}. For the generator network, the images are input to a $3-$layer convolutional network with $64\times128\times256$ feature maps, while the vectors are run through fully connected network with $256\times256$ hidden units with a final layer expands its size the same as the output of the convolutional network. The skill label is run through fully connected network with $128\times128$ hidden units with corresponding expansion to the size of input states. Then the input states and the skill label are now the same size and concatenated together, and are run through fully connected network with $256\times128$ hidden units to output actions and value. The selector network has the same architecture of the generator with input states and actions then outputs skill label. The discriminator network is similar to the selector, and we just adjust all hidden units to $64$ for the input states, actions and skill label, then run through fully connected network with $128\times64$ hidden units to output reward value. In the RTS task, the state information includes current resource of players and the images with $20\times20$ dimensions.
\subsection{Learning to Drive in Dense Traffic}
The experiment is conducted with the Next Generation Simulation program's Interstate 80 (NGSIM I-80) dataset\cite{halkias2006next}. NGSIM I-80 dataset includes various complex driver skills or behaviors such as lane changes, merges and sudden accelerations, and is recorded at different times of day with different traffic conditions, which contains uncongested and congested peak period. \cite{henaff2019model} provides an interactive simulation environment with this dataset, which is used in our driving task. 

We first apply preprocessing of the dataset for learning skill selection. The expert trajectories are labeled manually based on rules with three skills: $[1,0,0]$ corresponds to lane-change left, $[0,1,0]$ corresponds to lane keeping and $[0,0,1]$ corresponds to lane-change right. $150$ trajectories are sampled from demonstrations with the length of $13$s ($130$ frames). Each of modalities has $50$ trajectories.
\begin{table*}[t]
	\centering
	\setlength{\tabcolsep}{3.0mm}{
		\begin{tabular}{lrrrrr}  
			\toprule
			\multirow{2}{*}{Algorithms}  &  \multirow{2}{*}{\textit{Success Rate} (\%)} & \multirow{2}{*}{\textit{Mean Distance} (m)}& 
			\multicolumn{3}{c}{\textit{KL Divergence}}\\
			\cmidrule(lr){4-6}
			& & & Lane-change Left & Lane-keeping & Lane-change Right \\
			\midrule
			BC          &     $6.8\pm3.2$    &    $81.7\pm3.1$   &    $3827\pm358$    &   $4008\pm486$     &    $2581\pm371$  \\
			GAIL        &    $73.9\pm1.3$    &   $168.4\pm5.2$   &    $1764\pm279$    &   $1893\pm378$     &    $606\pm278$   \\
			CGAIL       &    $65.5\pm0.9$    &   $149.8\pm7.2$   &    $1297\pm255$    &   $1892\pm279$     &    $977\pm109$   \\
			Triple-GAIL &    $\mathbf{80.9\pm1.2}$    &   $\mathbf{179.6\pm3.6}$   &    $\mathbf{447\pm122}$     &   $\mathbf{685\pm214}$      &    $\mathbf{392\pm127}$   \\
			\midrule
			Expert        &         $100$      &    $210\pm2.1$    &          0         &        0           &       0          \\
			\bottomrule
		\end{tabular}
	}
	\caption{The \textit{Success Rate}, \textit{Mean Distance} and \textit{KL Divergence} of different algorithms.}
	\label{tab:ppuu}
\end{table*}
\begin{figure}[tb]
	\centering
	\subfigure[\small BC]{
		\begin{minipage}[t]{0.49\linewidth}     
			\centering
			\includegraphics[width=1.66in]{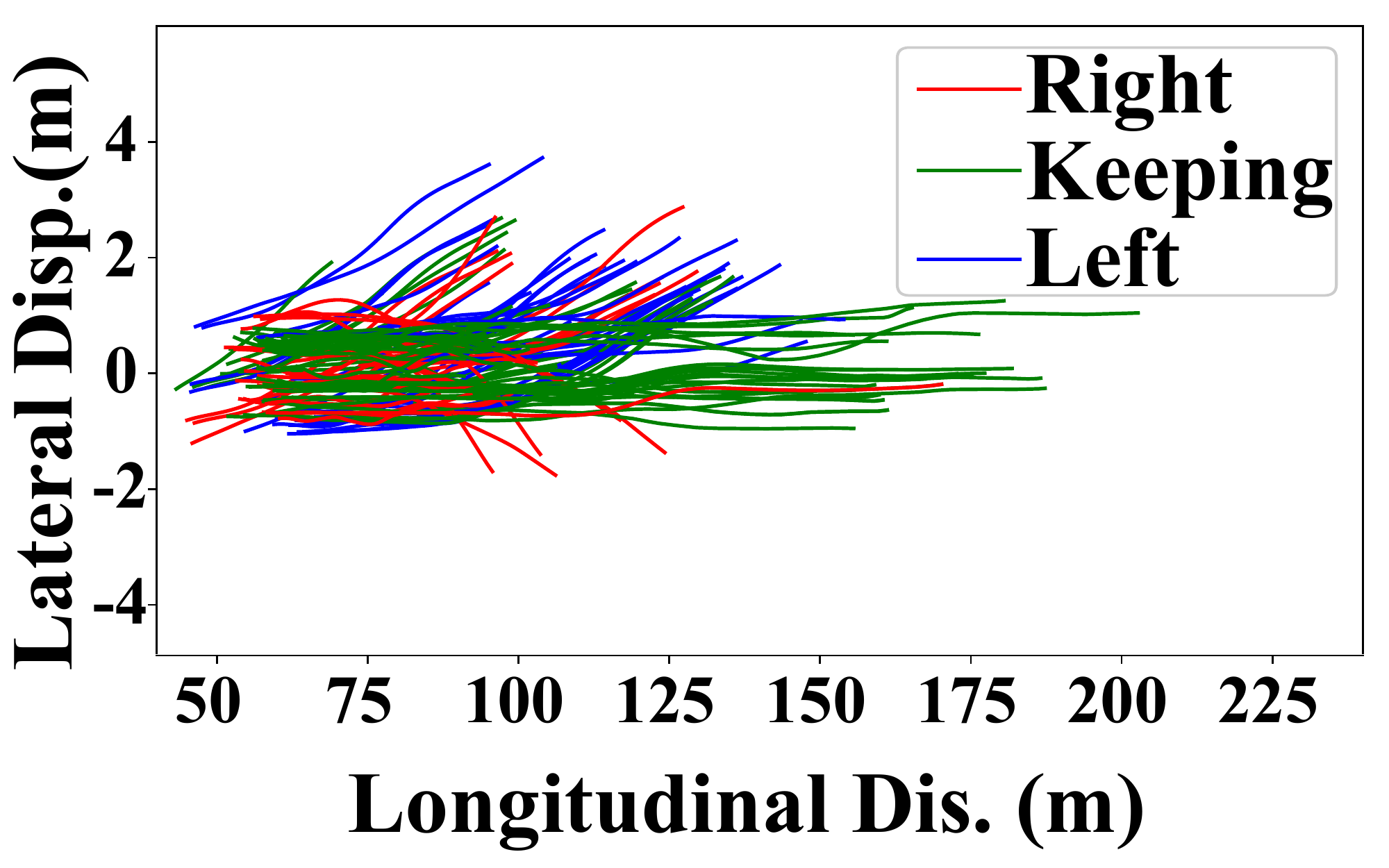}
			\label{fig:multi1}
		\end{minipage}%
	}%
	\subfigure[\small GAIL]{
		\begin{minipage}[t]{0.49\linewidth}
			\centering
			\includegraphics[width=1.66in]{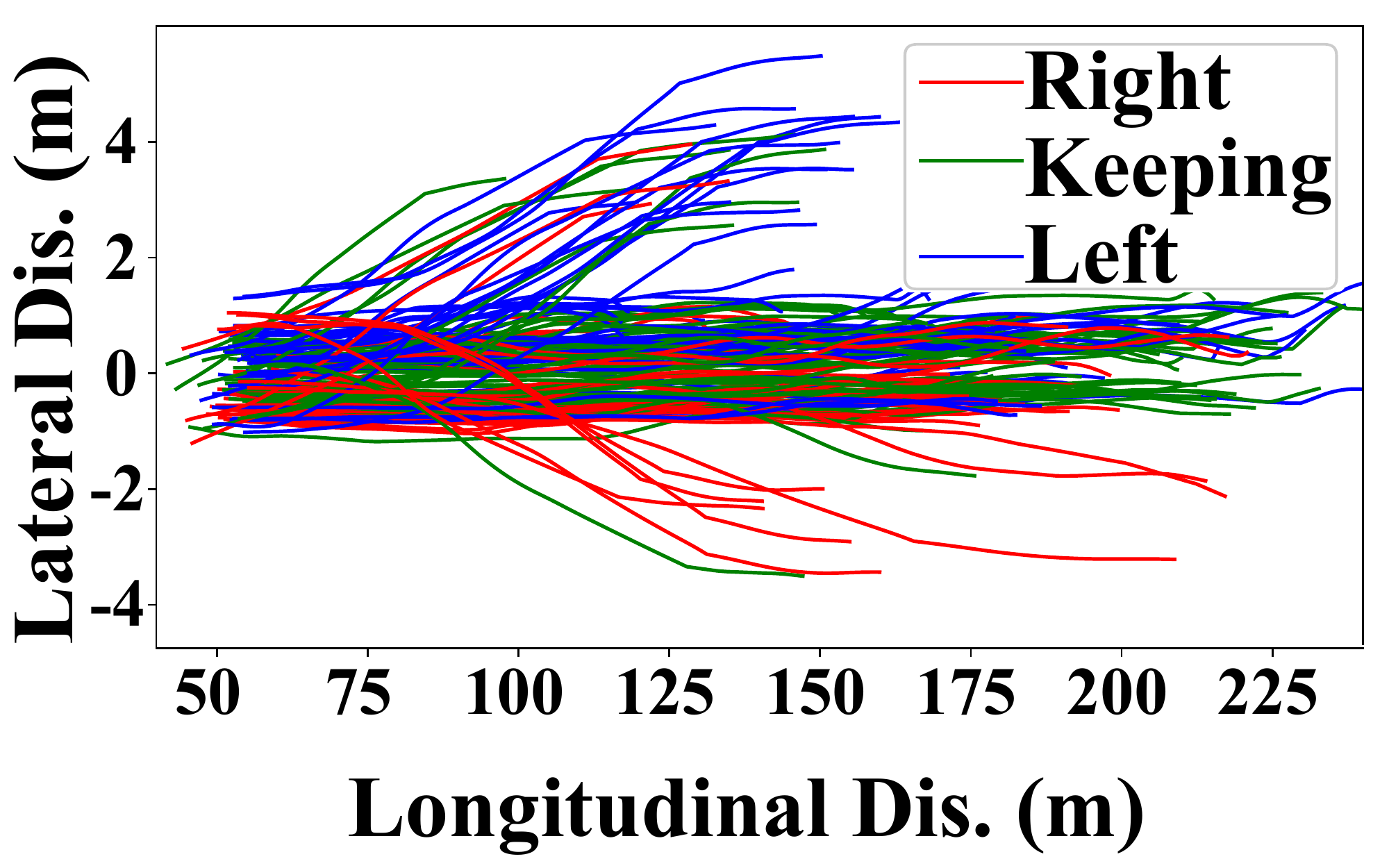}
			\label{fig:multi2}
		\end{minipage}
	}\\
	\subfigure[\small CGAIL]{
		\begin{minipage}[t]{0.49\linewidth}
			\centering
			\includegraphics[width=1.66in]{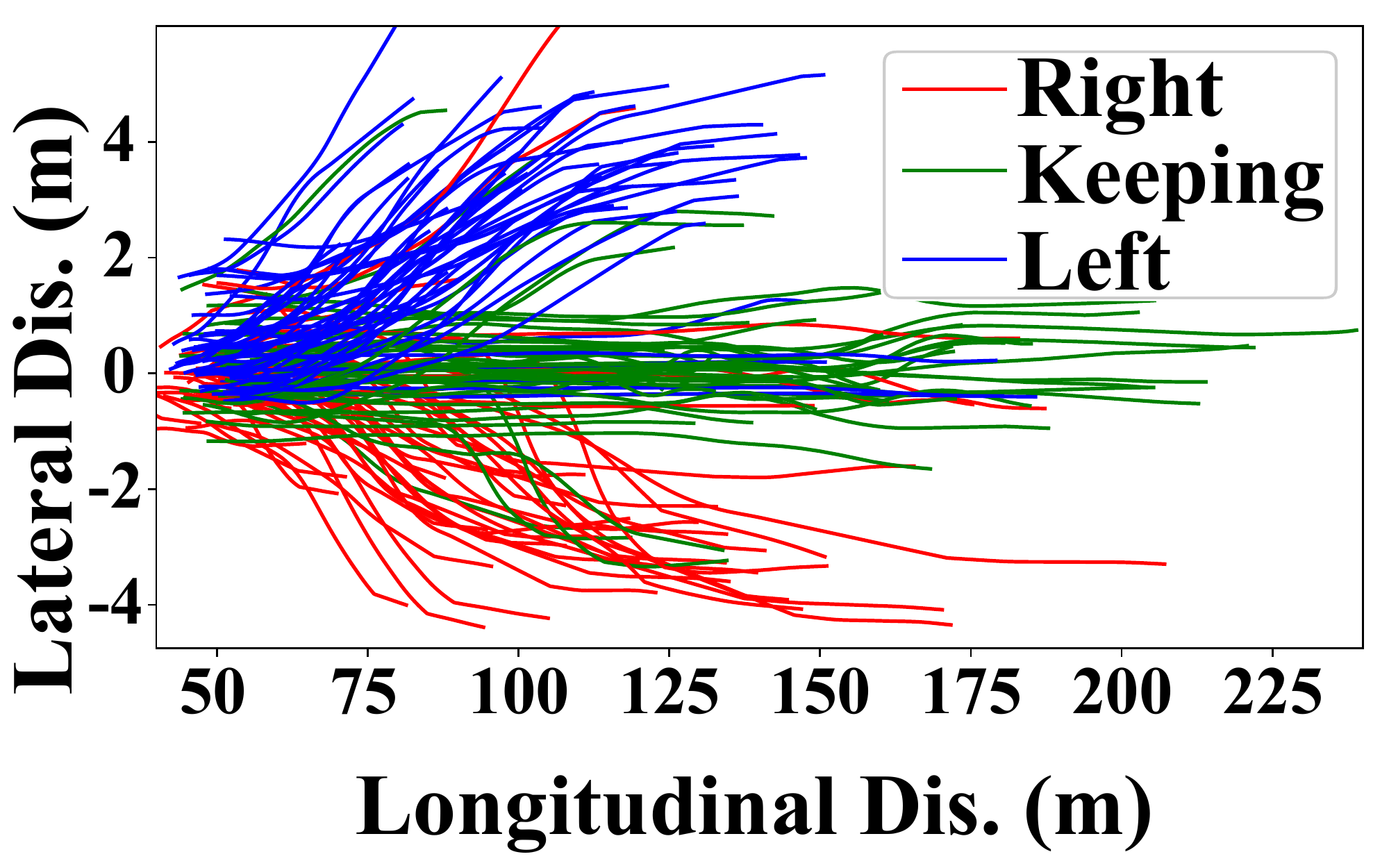}
			\label{fig:multi3}
		\end{minipage}
	}%
	\subfigure[\small Triple-GAIL]{
		\begin{minipage}[t]{0.49\linewidth}
			\centering
			\includegraphics[width=1.66in]{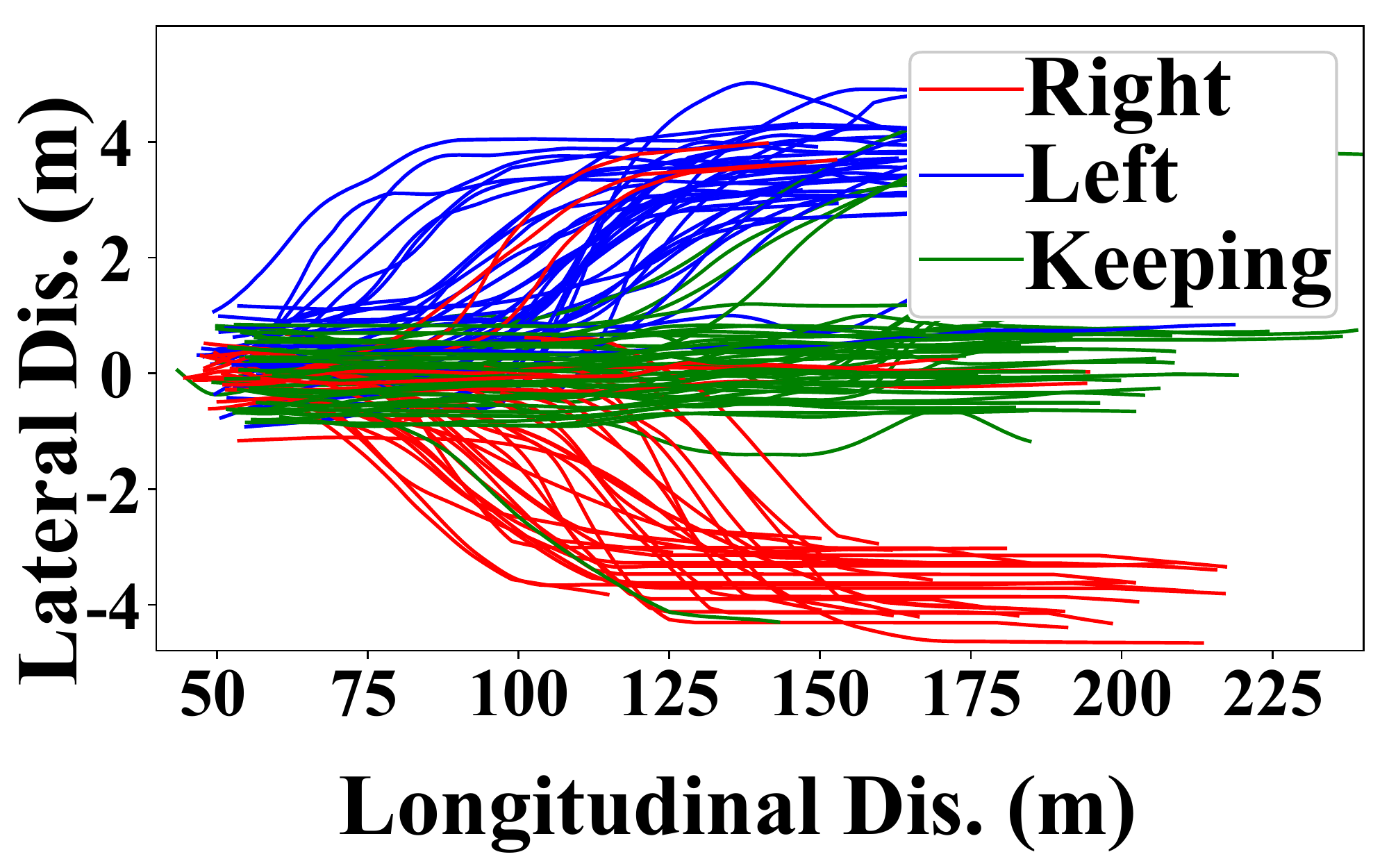}
			\label{fig:multi4}
		\end{minipage}
	}%
	\caption{The visualization of trajectories. The trajectories of lane-change right, lane keeping and lane-change left are represented by red, green and blue lines  respectively. Dis. denotes displacement.}
	\label{fig:multi}
\end{figure}
\subsubsection{Performance of Skill Selection}
We first estimate the performance of the selector, which acts on selecting driving skills. The rollouts are visualized in Figure \ref{fig:multi}, where the red, green and blue trajectories indicate various skills labeled by expert, corresponding to lane-change left, lane keeping and lane-change right respectively.

As shown in Figure \ref{fig:multi1}, the BC model has the shortest trajectory length because of collision. And the trajectories generated by BC tend to deviate to left side due to cascade errors. GAIL has longer trajectories, but fails to distinguish skill labels as shown in the mixed color of Figure \ref{fig:multi2}. We conclude that BC and GAIL have poor ability in distinguishing and selecting different skills from demonstrations due to the mode collapse problem.

Compared with BC and GAIL, CGAIL and Triple-GAIL have a more clear separation among rollout trajectories with different driving skills. However, CGAIL has relative short  trajectories, also indicated in \textit{Mean Distance} in Table \ref{tab:ppuu}. We guess that the pretrained classifier of CGAIL is unable to choose skill label adaptively guiding the generator for decision-making without joint optimization. In contrast, Triple-GAIL can successfully select driving skill as well as learn effective lane-change driving policies accordingly.

\begin{figure*}[tb]
	\centering
	\subfigure[\footnotesize S1]{
		\includegraphics[width=0.23\textwidth]{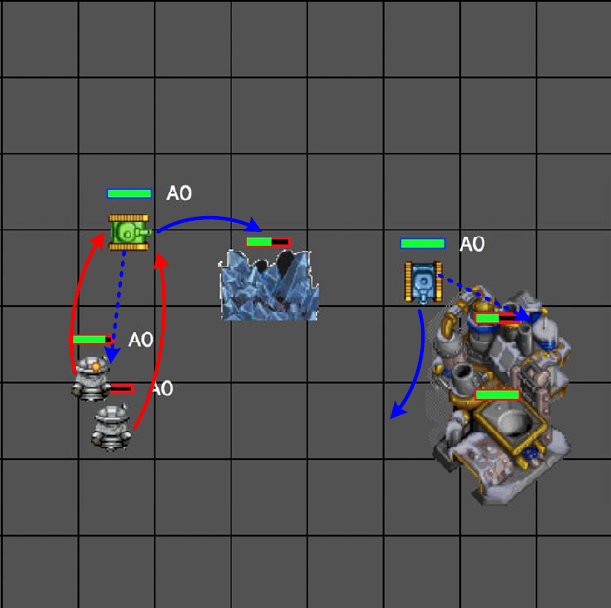}
		\label{fig:rts3}
	}%
	\subfigure[\footnotesize S2]{
		\includegraphics[width=0.23\textwidth]{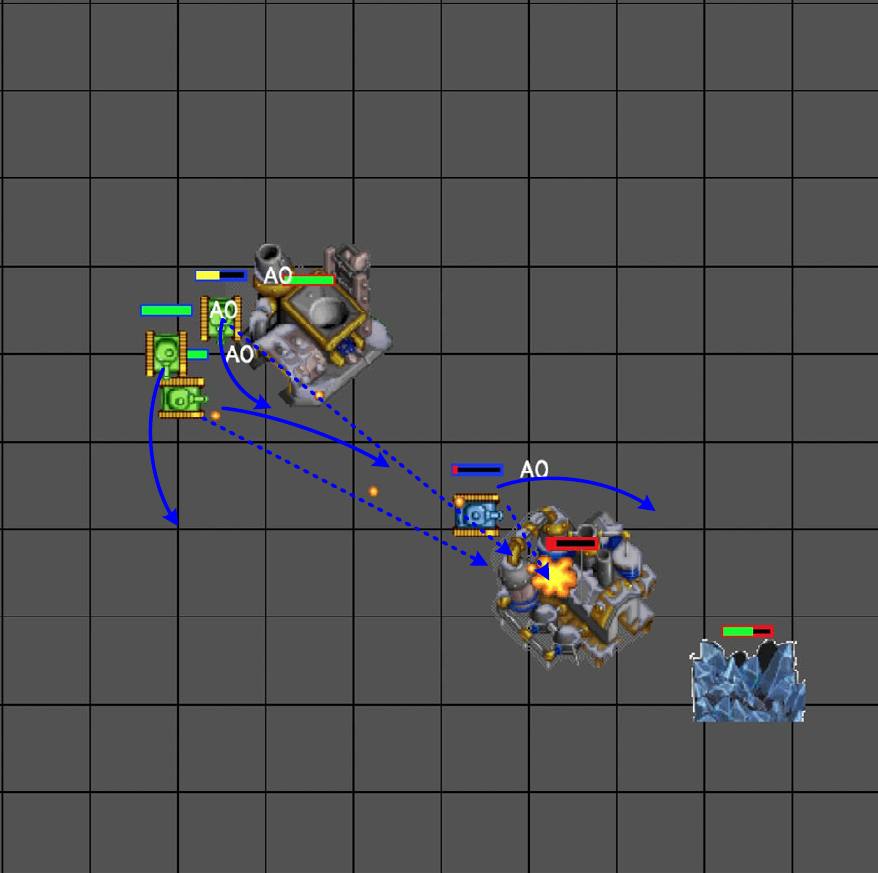}
		\label{fig:rts4}
	}%
	\subfigure[\footnotesize H1]{
		\includegraphics[width=0.23\textwidth]{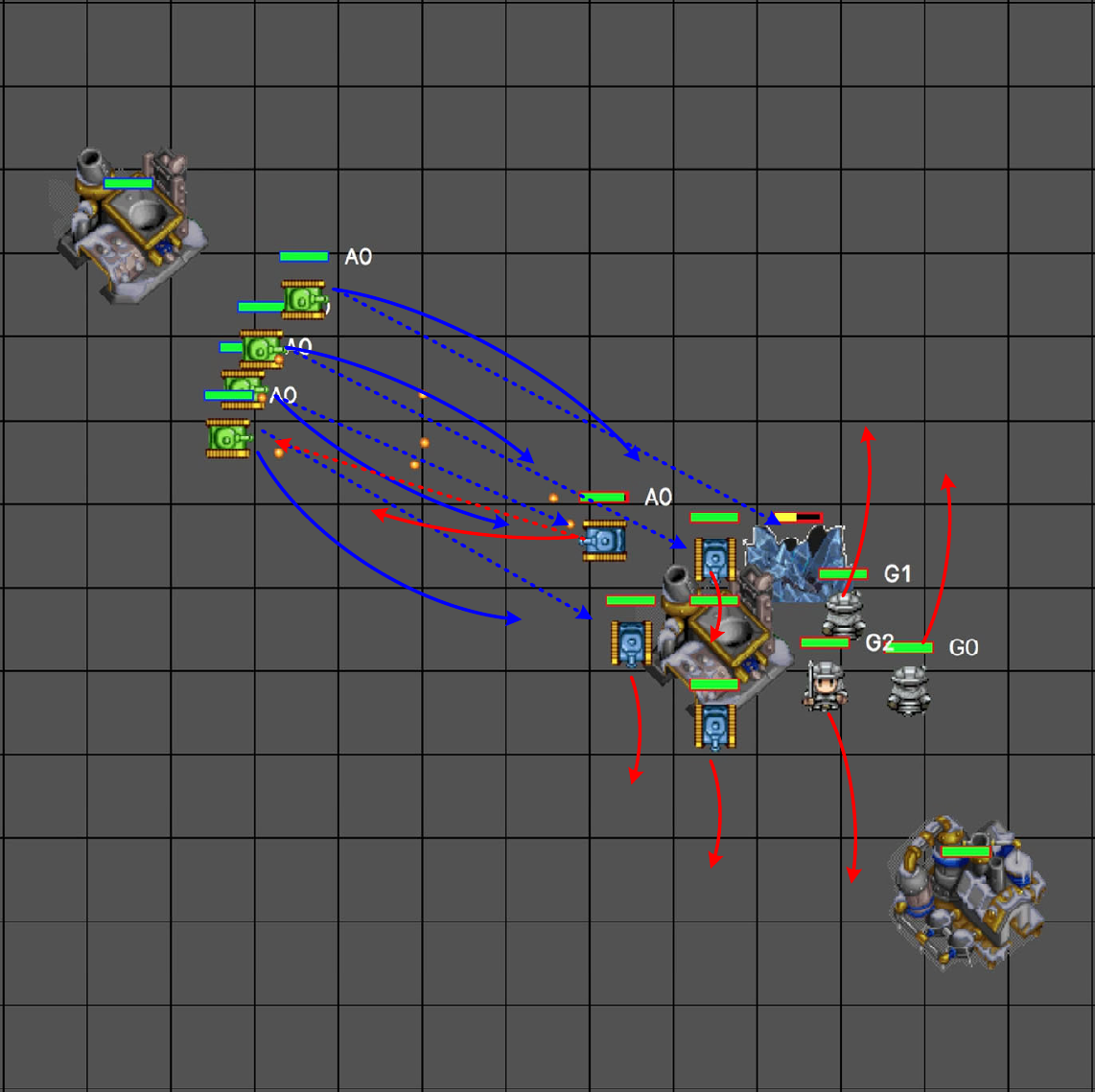}
		\label{fig:rts6}
		
	}%
	\subfigure[\footnotesize H2]{
		\includegraphics[width=0.23\textwidth]{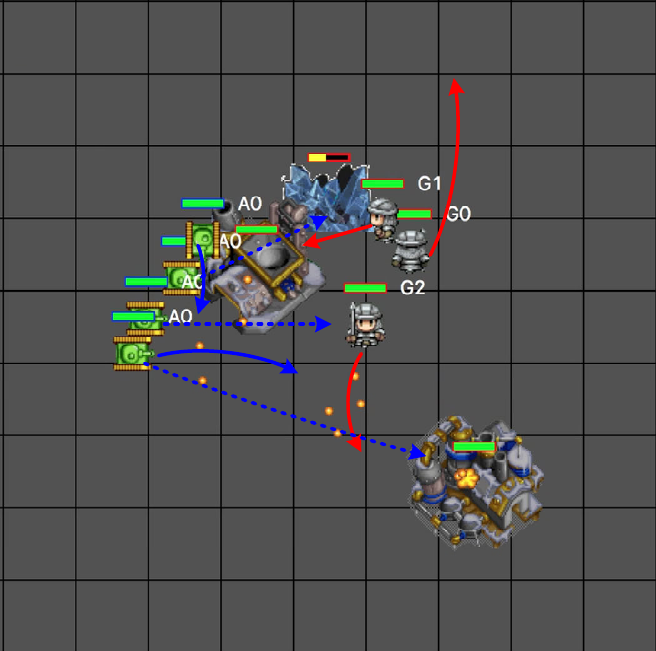}
		\label{fig:rts8}
		
	}%
	\caption{Game screenshots between trained agent (blue) and built-in agent (red). The agents' colors are shown on the arrows and the boundary of hit point gauges. (a)$\sim$(b) is an example case against \textit{SIMPLE} and (c)$\sim$(d) is an example case against \textit{HIT-N-RUN}. The solid line indicates the movements of agents while the dotted line points to the attack target.}
	\label{fig:rts}
\end{figure*}
\subsubsection{Overall Result of Policy Imitation}
We first evaluate our algorithm in two metrics: \textit{Success Rate} and \textit{Mean Distance}. The \textit{Success Rate} indicates the probability of moving through the road segment without collision, while the \textit{Mean Distance} indicates the distance travelled before the episode ends.
Then, in order to further evaluate the statistical distance between learned policy and expert policy, we sample the trajectory positions $(x,y)$ from demonstrations and rollouts separately, which can be regarded as the low-dimension projection of policies. 
The \textit{KL Divergence} of position distribution between generated data and demonstrations is also calculated. 

From Table \ref{tab:ppuu} we conclude that Triple-GAIL outperforms all other three baselines in both \textit{Success Rate} and \textit{Mean Distance}, and is closest to expert demonstrations. Furthermore, for different skills in this task including lane-change left, lane keeping and lane-change right, Triple-GAIL indicates the smallest \textit{KL Divergence} compared with other baselines, which means with appropriate skill selection, Triple-GAIL is able to imitate corresponding expert policies effectively.
\begin{table}
	\centering
	\setlength{\tabcolsep}{10.0mm}{
		\begin{tabular}{lr}  
			\toprule
			Algorithms                 &   Accuracy     \\
			\midrule  
			CGAIL                      &    $83.2\%$    \\
			Triple-GAIL                &    $\mathbf{90.7\%}$    \\
			Triple-GAIL$\backslash R_E$  &    $69.6\%$    \\
			Triple-GAIL$\backslash R_G$  &    $81.3\%$    \\
			\bottomrule
		\end{tabular}
	}
	\caption{Selection accuracies of driving skills. Triple-GAIL$\backslash R_E$ removes supervised loss $R_E$ while Triple-GAIL$\backslash R_G$ removes supervised loss $R_G$.}
	\label{tab:acc}
\end{table}
\subsubsection{Ablation Study}
To further estimate the performance of the selector and the joint optimization of the selector and the generator, the selection accuracies of driving skills is compared in  Table \ref{tab:acc}. We show that both CGAIL and Triple-GAIL have high selection accuracies, while Triple-GAIL is slightly higher up to $90\%$. The comparison between CGAIL and Triple-GAIL illustrates that the joint optimization of the selector and the generator in Triple-GAIL is superior to CGAIL which is pretrained and fixed. There is a significant difference in Triple-GAIL with and without the cross-entropy loss term $R_E$, which shows that the supervised signal from demonstrations plays an important role in training the selector. In addition, the comparison of Triple-GAIL with and without the $R_G$ also validates the advantage of $R_G$.

\begin{table}
	\centering
	\setlength{\tabcolsep}{4.0mm}{
		\begin{tabular}{lrr}  
			\toprule
			Algorithms  &  \textit{SIMPLE}   &  \textit{HIT-N-RUN} \\
			\midrule  
			BC          &    $42.7\pm5.6$    &    $31.5\pm3.8$     \\
			GAIL        &    $60.4\pm7.2$    &    $52.4\pm6.4$     \\
			CGAIL       &    $67.5\pm5.6$    &    $60.5\pm6.3$     \\
			Triple-GAIL  &    $\mathbf{73.9\pm3.2}$    &    $\mathbf{68.3\pm5.6}$     \\
			\midrule
			CGAIL$+label$       &    $69.5\pm6.9$    &    $62.7\pm7.3$     \\
			Triple-GAIL$+label$  &    $78.9\pm3.2$    &    $76.3\pm6.6$     \\
			\midrule
			Expert Matched     &    $95.2\pm1.7$    &    $90.6\pm4.9$     \\
			Expert Mismatched  &    $38.2\pm4.5$    &    $54.3\pm5.9$     \\
			\bottomrule
		\end{tabular}
	}
	\caption{Win rates of different algorithms competing with built-in agents over $10$k games. \textit{+label} denotes that the inferred labels are replaced by true labels. Matched denotes the agents run against the targeted built-in agents while Mismatched denotes the agents run against the mismatched built-in agents.}
	\label{tab:rts}
\end{table}
\subsection{Learning to Play RTS Game}
We then verify the Triple-GAIL in a Mini-RTS game, which is a miniature version of StarCraft \cite{tian2017elf}. In Mini-RTS game, players are required to gather resources, build troops and finally invade/defend the enemy until one player wins. 
There are two built-in agents in Mini-RTS: \textit{SIMPLE} and \textit{HIT-N-RUN}. \textit{SIMPLE} is a conservative strategy, where all troops stay on the defensive until the number of melee tanks reaches up to $5$. After that, all troops launch a counterattack. \textit{HIT-N-RUN} usually builds $2$ range tanks that move towards enemy base to harass the opponent, taking advantage of long attack range and high speed. We train the agent with frame-skip of $50$ and history length of $20$. Other settings and details refer to \cite{tian2017elf}.  

In order to gather demonstrations with multi-modal policies, two sets of targeted instructions adopted for the above two built-in agents respectively. The win rates of these two targeted instructions reach up to $90\%$ while less than $60\%$ if the agents run against the mismatched built-in agents, as shown in Table \ref{tab:rts}. Then the demonstrations with multiple policies are sampled by running the corresponding games and labeling the sampled state-action pairs. After that, Triple-GAIL and three baselines are trained.

We compare the performance of all four algorithms and the win rates are listed in Table \ref{tab:rts}. It is clear that Triple-GAIL provides better performance than all other baselines in both two built-in agents. We demonstrate that Triple-GAIL can effectively distinguish the enemy's tactical intention and helps to adopt the corresponding policy. When specifying skill labels with expert demonstration (CGAIL$+label$) instead of inferring from the classifier of CGAIL, we confirm that the joint optimization of the selector and the generator in Triple-GAIL indeed improves the policy imitation performance.

Figure \ref{fig:rts} provides typical game screenshots of Triple-GAIL. When the trained agents versus \textit{SIMPLE}, the learned agent commonly builds one range tank and directly moves towards to enemy base and attack energy troops in range, continuing to harass the enemy, as shown in the top row of Figure \ref{fig:rts3}. Once gaining an advantage, all melee and range tanks launch a general attack as shown in Figure \ref{fig:rts4}; if the opponent is \textit{HIT-N-RUN}, the trained agent will firstly build several range tanks against the harassment as shown in Figure \ref{fig:rts6}, and immediately launch a counterattack if enemy is weakly guard as shown in the Figure \ref{fig:rts8}. 

\section{CONCLUSION}
In this paper, we propose Triple-GAIL, a novel multi-modal GAIL framework that is able to learn skill selection and imitation jointly from both expert demonstrations and continuously generated experiences by introducing an auxiliary selector. We provide theoretical guarantees on the convergence to optima for both of the generator and the selector respectively. Experiments on driving task and real-time strategy game demonstrate that Triple-GAIL can better fit multi-modal behaviors close to the demonstrators and outperforms state-of-the-art methods. 

\bibliographystyle{IEEEtran}
\bibliography{root}

\end{document}